\documentclass{article}

\usepackage[margin=1in]{geometry}
\usepackage[utf8]{inputenc}
\usepackage[T1]{fontenc}
\usepackage{hyperref}
\usepackage{url}
\usepackage{booktabs}
\usepackage{amsmath}
\usepackage{amssymb}
\usepackage{graphicx}
\usepackage{enumitem}
\usepackage{natbib}
\usepackage{xcolor}

\title{Do LLMs Know What They Know? \\Measuring Metacognitive Efficiency with Signal Detection Theory}

\author{%
  Jon-Paul Cacioli \\
  Independent Researcher \\
  Melbourne, Australia \\
  \texttt{synthium@hotmail.com} \\
  ORCID: 0009-0000-7054-2014
}

\begin{document}

\maketitle

\begin{abstract}
Standard evaluation of LLM confidence relies on calibration metrics (ECE, Brier score) that conflate two distinct capacities: how much a model knows (Type-1 accuracy) and how well its confidence signal tracks that knowledge (Type-2 metacognitive sensitivity). We apply Signal Detection Theory (SDT) to decompose these capacities in large language models, treating token-level normalised log-probability as a graded confidence variable and answer correctness as the state to be discriminated. We characterise the Type-2 ROC of this signal, including its unequal-variance structure via z-ROC analysis, and---because the meta-$d'$ efficiency ratio is not well defined for open-ended QA, which lacks a two-alternative Type-1 decision---quantify metacognitive efficiency with a model-free information measure, normalised metacognitive information (meta-$I_{2r}$). Applied to four LLMs (Llama-3-8B-Instruct, Mistral-7B-Instruct-v0.3, Llama-3-8B-Base, Gemma-2-9B-Instruct) across 224{,}000 factual QA trials, we find: (1)~metacognitive information varies by a factor of 1.98 across models and is not predicted by accuracy---the rank correlation is ${-}0.80$ on TriviaQA and ${+}0.00$ on Natural Questions, and Gemma-2 is reliably the least informative on both; (2)~the confidence signal has model-specific unequal-variance structure (z-ROC slopes from 0.78 to 1.18) that is invisible to calibration metrics, with the slope ordering replicating on Natural Questions; (3)~metacognitive information is domain-specific, weakest in Science \& Technology for every model; (4)~temperature dissociates Type-1 accuracy from metacognitive information, which stays near-flat for three of four models while accuracy falls; and (5)~metacognitive information tracks the accuracy gain from confidence-based abstention exactly ($\rho={+}1.00$) while accuracy does not. All estimates carry permutation nulls and bootstrap confidence intervals. \textbf{This version (v3) corrects a differential length bias in the automated correctness scorer; the inverse accuracy--efficiency coupling reported in v1 and v2 does not survive that correction. See the version note.} Pre-registered analysis; code and data publicly available.
\end{abstract}

\footnotetext[1]{\textbf{Version note (v3).} The correctness labels used in
v1 and v2 were produced by an automated scorer (exact match against verified
aliases, then a string-similarity threshold) that has since been found to be
\emph{differentially} length-biased. Validated against 1{,}200 human
adjudications, the scorer missed 53\% of one model's human-correct answers and
26\% of another's, because it fails on answers that contain the correct alias
inside a longer sentence and the models differ substantially in verbosity. A
containment fallback --- accepting a match when the normalised token sequence of
any alias appears contiguously in the normalised answer --- raises agreement
with human labels from 0.76 to 0.86 at a false-positive cost of 1.4\%
(adjudicated on a further 630 blinded responses). All analyses in this version
are recomputed on the corrected labels.

\emph{What changes.} Accuracy rises for every model, by between 2.7 and 13.3
points, and the accuracy ordering changes. The cross-model efficiency ordering
does not survive: Mistral moves from the highest meta-$I_{2r}$ to the third of
four, and the inverse coupling with accuracy weakens from $\rho={-}1.00$ to
$\rho={-}0.80$ on TriviaQA and disappears on Natural Questions
($\rho={+}0.00$). The two-fold cross-model range becomes 1.98-fold. Section~4.3
is rewritten accordingly. \emph{What survives.} The Type-2 SDT structure and
its slope ordering, replicating on NQ; the NLP monotonicity and z-ROC linearity
checks; domain-specificity, with Science \& Technology weakest for all four
models; the temperature dissociation; and the selective-prediction result,
which strengthens ($\rho={+}1.00$ between meta-$I_{2r}$ and abstention gain,
against ${+}0.80$ in v2). Residual differential scoring error remains, so the
corrected values should be treated as better than v2's rather than as final.

An earlier version of this paper estimated metacognitive efficiency using meta-$d'$/$M$-ratio. That estimator requires a two-alternative Type-1 detection decision, which open-ended factual QA does not provide; forcing the mapping makes $d'$ and meta-$d'$ functions of the same correctness-by-confidence table, so $M$-ratio is pinned near 1 by construction and reported cross-model $M$-ratio differences reflect departures from the equal-variance assumption rather than metacognitive efficiency. This version replaces $M$-ratio throughout with a model-free measure (meta-$I_{2r}$) and reports the Type-2 SDT structure (AUROC$_2$, z-ROC slope, $d_a$) directly. All analyses were re-run on the same 224{,}000-trial dataset. The domain-specificity and temperature-dissociation findings are unchanged; the direction of the cross-model efficiency finding reverses under the corrected measure.}

\section{Introduction}

When a large language model (LLM) answers a factual question, two capacities determine the reliability of its output: its ability to discriminate correct from incorrect responses, and its ability to \emph{monitor} that discrimination through its confidence signal. These are fundamentally different problems requiring different interventions. A model that cannot discriminate needs better training data or architectural improvements. A model that discriminates well but monitors poorly needs recalibration, not retraining. Current evaluation practice does not make this distinction.

Consider two hypothetical models evaluated on the same factual QA benchmark. Model~A reports 90\% confidence on every trial and achieves 90\% accuracy; its Expected Calibration Error (ECE) is near zero. Model~B reports 95\% confidence when correct and 60\% when incorrect, but its average confidence overshoots its 80\% accuracy; its ECE is worse than Model~A's. Yet Model~B's confidence is \emph{far more useful}: it tells you which specific answers to trust. Model~A's confidence, despite perfect calibration, carries no information about correctness. The standard ECE metric rewards the wrong model.

This example illustrates a well-known limitation of ECE~\citep{guo2017calibration}: it measures the average alignment between confidence and accuracy, conflating the \emph{resolution} of the confidence signal (how well it separates correct from incorrect) with its \emph{bias} (the overall level of confidence). The Brier score decomposes into reliability, resolution, and uncertainty, but not into the model's discriminative capacity and its metacognitive sensitivity controlling for that capacity. The Area Under the Type-2 ROC (AUROC$_2$) of the confidence--accuracy curve~\citep{steyvers2025metacognition} improves on ECE by measuring ranking quality. It does not normalise for the base rate of correct responses; the information-theoretic measure we adopt does, though we note (\S\ref{sec:results}) that on a sample whose accuracies yield near-identical base-rate entropies the normalisation is numerically inert and the two measures coincide.

Signal Detection Theory (SDT; \citealt{green1966signal}; \citealt{macmillan2005detection}) provides the natural framework for this decomposition. Developed over seven decades of psychophysical research, SDT separates performance into \emph{sensitivity} (how well an observer discriminates two states) and \emph{criterion} (the observer's threshold for responding). \citet{cacioli2026llms} demonstrated that the parametric SDT framework---ROC analysis, unequal-variance model fitting, criterion estimation---reveals structure in LLM confidence invisible to calibration metrics alone.

The present work applies SDT at the \emph{Type-2} level: how well does a model's confidence signal discriminate its own correct from incorrect answers? We treat token-level normalised log-probability (NLP) as a graded confidence variable and characterise the Type-2 ROC it produces, including its unequal-variance structure. For the \emph{efficiency} question---how much of the available correctness information the confidence signal captures---we use a model-free information-theoretic measure rather than the meta-$d'$ ratio, for reasons we make explicit in \S\ref{sec:whynotmetad}: open-ended QA has no two-alternative Type-1 decision, so meta-$d'$ is not well defined without reintroducing assumptions the data do not support.

We quantify metacognitive efficiency using the mutual-information approach of \citet{dayan2023metacognitive}, who introduced meta-$I$ (the mutual information between choice accuracy and confidence) and normalised variants that quantify efficiency. We use the accuracy-entropy normalisation, which we denote meta-$I_{2r}$:\footnote{Notation ours. \citet{dayan2023metacognitive} defines meta-$I$ and several normalisations; meta-$I_{2r}$ here refers specifically to the ratio of transmitted information to accuracy entropy in Eq.~\ref{eq:metai2r}. \citet{fitousi2025information} empirically validates this family of information-theoretic efficiency measures.}
\begin{equation}
\text{meta-}I_{2r} = \frac{I(\text{correct}\,;\,\text{confidence})}{H(\text{correct})},
\label{eq:metai2r}
\end{equation}
the mutual information between binary correctness and the discretised confidence variable, normalised by the entropy of accuracy. meta-$I_{2r}$ is 0 when confidence is uninformative about correctness and approaches its ceiling when confidence fully resolves the correct/incorrect distinction. It is model-free: it assumes no Gaussian evidence distribution, requires no Type-1 decision, and is defined identically for two-way, $n$-way, and open-ended tasks. Because plug-in mutual information is upward biased at finite sample sizes, every estimate is bias-corrected against a permutation null and reported with a bootstrap confidence interval.

Our contributions are:
\begin{enumerate}[leftmargin=*, nosep]
    \item We give a Type-2 SDT characterisation of the LLM confidence signal---AUROC$_2$, z-ROC slope, and unequal-variance sensitivity $d_a$---and show the signal has model-specific variance structure invisible to ECE.
    \item We introduce meta-$I_{2r}$ as a model-free metacognitive-efficiency measure for LLM confidence, appropriate where meta-$d'$ is undefined, and show it reveals a model that has the lowest accuracy yet the highest metacognitive information.
    \item We show metacognitive efficiency is domain-specific, and that temperature dissociates Type-1 accuracy from metacognitive information.
    \item All analyses are pre-registered, with permutation nulls, bootstrap CIs, and a Natural Questions replication; code and data are public.
\end{enumerate}

The contribution is an evaluation methodology, not a dataset or benchmark. Limitations of scope---four open-weight 7--9B models, two factual QA datasets, quantised inference---are detailed in \S\ref{sec:limitations}.

\section{Background: Type-2 Signal Detection Theory}

\subsection{The confidence signal as a Type-2 detector}

In the Type-1 SDT framework applied to LLM factual QA~\citep{cacioli2026llms}, each question is a trial in which the model generates an answer. The normalised log-probability (NLP) of the generated answer serves as the evidence variable: $\text{NLP} = (1/L) \sum_{i=1}^{L} \log p(t_i \mid t_{<i})$, where $L$ is the answer length in tokens. Higher NLP indicates greater model confidence.

The Type-2 question~\citep{galvin2003type} is how well this confidence signal discriminates the model's own correct from incorrect responses. Given the binary correctness of each trial and the graded NLP, the Type-2 ROC plots the hit rate (proportion of correct answers above a confidence criterion) against the false-alarm rate (proportion of incorrect answers above it), swept across criteria. The area under this curve, AUROC$_2$, is a non-parametric measure of how well confidence separates correct from incorrect answers, and requires no distributional assumptions.

\subsection{Unequal-variance structure via z-ROC}
\label{sec:zroc}

The \emph{shape} of the Type-2 ROC carries information beyond its area. Under the Gaussian SDT model, plotting the ROC in $z$-coordinates (probit-transformed hit and false-alarm rates) yields a straight line whose slope $s$ equals the ratio of the standard deviations of the two underlying evidence distributions~\citep{green1966signal, macmillan2005detection}. A slope $s=1$ indicates equal variance; $s<1$ indicates the correct-answer evidence distribution is more variable than the incorrect-answer distribution, and $s>1$ the reverse. The unequal-variance sensitivity index $d_a = \sqrt{2/(1+s^2)}\,\cdot\,(\text{z-intercept})$ summarises sensitivity while respecting this asymmetry. These are properties of the empirical ROC, estimated directly by regression on the z-ROC, and involve no ideal-observer inversion.

\subsection{Why not meta-$d'$?}
\label{sec:whynotmetad}

The standard Type-2 efficiency measure is meta-$d'$~\citep{maniscalco2012signal, maniscalco2014signal}, which asks what Type-1 sensitivity an ideal observer would need to reproduce an observed pattern of confidence ratings, normalised as the ratio $M = \text{meta-}d'/d'$~\citep{fleming2014measure}. meta-$d'$ is defined for a two-alternative detection task: it requires a Type-1 decision (respond ``S1'' vs ``S2'') whose sensitivity $d'$ supplies the denominator, and a confidence report layered on that decision. Free-form factual QA has neither. There is no signal-absent trial and no binary Type-1 response; the model always emits an answer, and NLP is the only graded internal signal. Forcing the mapping---treating correctness as the S1/S2 variable and confidence bins as the rating---makes the estimated $d'$ and meta-$d'$ functions of the \emph{same} correctness-by-confidence table, so $M$ is pinned near 1 by construction and reflects only departures from the equal-variance assumption rather than metacognitive efficiency. This is not a limitation of SDT but of one Type-2 estimator applied outside its domain; recent measurement work reaches the same conclusion for $n$-choice tasks generally~\citep{rahnev2025measuring}. Our own data make the point concrete: the $M$-ratio values a naive application produces are rank-ordered almost perfectly by the z-ROC slope $s$ of the same models (Spearman $\rho = +0.95$; \S\ref{sec:zroc}, Table~\ref{tab:sdt}), confirming that in this paradigm $M$ tracks the unequal-variance structure of the confidence ROC and not a metacognitive-efficiency quantity. We therefore report the Type-2 ROC structure directly (\S\ref{sec:zroc}) and use a model-free efficiency measure (\S\ref{sec:whyinfo}).

\subsection{Metacognitive information as a model-free efficiency measure}
\label{sec:whyinfo}

Normalised metacognitive information (Eq.~\ref{eq:metai2r}; \citealt{dayan2023metacognitive}) measures how much the confidence signal reduces uncertainty about correctness, as a fraction of the total uncertainty in correctness. It is model-free and defined for any task, and has been validated as a metacognition measure with explicit multi-alternative applicability~\citep{fitousi2025information}. We report it with a permutation null (shuffling confidence against correctness) to control the upward bias of plug-in mutual information, and with trial-level bootstrap confidence intervals.

\subsection{Why NLP is a valid confidence variable}

Verbalised confidence---prompting a model to report a numerical certainty score---is the dominant paradigm for LLM uncertainty estimation in black-box settings~\citep{xiong2024can}. However, \citet{dai2026rescaling} demonstrate that verbalised confidence suffers from severe discretisation: more than 78\% of responses on a 0--100 scale concentrate on three round-number values, producing sparse and unreliable estimates. Token-level log-probabilities, by contrast, provide a continuous confidence variable that is a direct output of the model's generative process. NLP is not a pure ``metacognitive signal'' in any cognitive sense; it is a fluency measure that reflects both the quality of the generated answer and the model's distributional properties~\citep{cacioli2026llms}. We adopt a \emph{functional} operationalisation: metacognitive monitoring is the discriminability of an internal signal for correctness, without requiring a distinct second-order system. This parallels the use of Type-2 measures in animal metacognition research~\citep{smith2014metacognitive}. As an empirical validation, we verify that NLP is monotonically related to accuracy across all conditions (\S\ref{sec:results}, Appendix~\ref{app:monotonicity}).

\section{Method}

\subsection{Models and Data}

Four LLMs spanning three model families were evaluated: Llama-3-8B-Instruct and Llama-3-8B-Base (Meta; \citealt{meta2024llama3}), Mistral-7B-Instruct-v0.3~\citep{jiang2023mistral}, and Gemma-2-9B-Instruct (Google; \citealt{team2024gemma2}). All were run as Q5\_K\_M quantisations via \texttt{llama-cpp-python}~0.3.16 with Vulkan backend on an AMD RX~7900~GRE (16\,GB VRAM).

\begin{table}[h]
\centering
\caption{Model summary. All models run as Q5\_K\_M GGUF quantisations.}
\label{tab:models}
\small
\begin{tabular}{llccc}
\toprule
\textbf{Model} & \textbf{Family} & \textbf{Params} & \textbf{Instruct} & \textbf{Quant size} \\
\midrule
Llama-3-8B-Instruct & Meta & 8B & Yes & 5.7\,GB \\
Llama-3-8B-Base & Meta & 8B & No & 5.7\,GB \\
Mistral-7B-Instruct-v0.3 & Mistral AI & 7B & Yes & 5.1\,GB \\
Gemma-2-9B-Instruct & Google & 9B & Yes & 6.7\,GB \\
\bottomrule
\end{tabular}
\end{table}

Two factual question-answering datasets were used. \textbf{TriviaQA}~\citep{joshi2017triviaqa}: 5{,}000 questions from the unfiltered set (seed\,=\,42), classified into four knowledge domains: History~\&~Politics (1{,}248), Arts~\&~Literature (1{,}167), Geography (667), Science~\&~Technology (634), plus Unclassified (1{,}284; excluded from domain analyses). \textbf{Natural Questions}~\citep{kwiatkowski2019natural}: 3{,}000 short-answer questions from NQ-Open, a replication dataset. Each model answered each question at seven temperatures $T \in \{0.1, 0.3, 0.5, 0.7, 1.0, 1.5, 2.0\}$, yielding 224{,}000 trials. Per trial we recorded the generated answer, NLP, and binary correctness (exact match against verified aliases, \texttt{difflib.SequenceMatcher} $\geq 0.85$ fallback). Data for the three original models were collected under a prior pre-registration~\citep{cacioli2026llms}; Gemma-2 was added post-registration following the identical protocol.

\subsection{Pipeline}
\label{sec:pipeline}

\paragraph{Confidence binning.} NLP values are binned into $2K$ ordered categories ($K=4$), with edges at the $\{12.5,\ldots,87.5\}$th quantiles of the NLP distribution at $T=1.0$ within each model~$\times$~dataset condition, held constant across temperatures.

\paragraph{Measures.} For each analysis cell we compute (i) AUROC$_2$; (ii) the z-ROC slope $s$ and $d_a$ by linear regression on the probit-transformed empirical Type-2 ROC (\S\ref{sec:zroc}); and (iii) meta-$I_{2r}$ (Eq.~\ref{eq:metai2r}), bias-corrected against a permutation null of 2{,}000 confidence--correctness shuffles.

\paragraph{Inference.} All confidence intervals are 95\% bootstrap percentile intervals from 2{,}000 trial-level resamples (seed\,=\,42); each resample recomputes the full pipeline.

\subsection{Pre-Registered Hypotheses}

Analyses for the three original models were pre-registered (OSF: \url{https://osf.io/5q7mt}); Gemma-2 is a post-registration generalisability test. Hypotheses are tested at $T=1.0$ on TriviaQA. \textbf{H1} (cross-model variation): metacognitive efficiency varies across models. \textbf{H2} (domain-specificity): efficiency varies across TriviaQA domains. \textbf{H3} (temperature dissociation): metacognitive efficiency is stable while Type-1 accuracy varies across $T \in \{0.3,0.5,0.7,1.0\}$. \textbf{H4} (efficiency not determined by accuracy): the ranking of models by metacognitive efficiency differs from their ranking by accuracy. (The pre-registration specified these hypotheses in terms of meta-$d'$/$M$-ratio; we test the identical conceptual claims with meta-$I_{2r}$, for the reasons in \S\ref{sec:whynotmetad}.)

\section{Results}
\label{sec:results}

\subsection{Validation}
All eight model$\times$dataset conditions at $T{=}1.0$ passed the NLP monotonicity check: accuracy increased strictly across NLP quartiles (Appendix~\ref{app:monotonicity}). All z-ROC fits were highly linear ($R^2 \geq 0.98$), supporting the Gaussian SDT model for the confidence signal.

\subsection{SDT structure of the confidence signal (H1, structure)}

\begin{table}[t]
\centering
\caption{SDT structure of the Type-2 (correct/incorrect) confidence ROC at $T{=}1.0$, \textbf{corrected labels}. $s$ is the z-ROC slope (ratio of incorrect-to-correct evidence SD); $s<1$ indicates the correct-answer evidence distribution is more variable. $d_a$ is the unequal-variance sensitivity index. Slopes from linear regression on the empirical z-ROC ($R^2 = 0.988$--$1.000$ across cells); the ordering of $s$ replicates on NQ. Values in v2 are given in parentheses.}
\label{tab:sdt}
\small
\begin{tabular}{lcccc}
\toprule
\textbf{Model} & \textbf{AUROC$_2$} & \textbf{z-ROC slope $s$} & $\boldsymbol{d_a}$ & \textbf{$s$ (NQ)} \\
\midrule
Llama-3-Instruct & 0.833 \emph{(0.831)} & 0.905 \emph{(0.863)} & 1.354 \emph{(1.385)} & 0.936 \emph{(0.852)} \\
Mistral-Instruct & 0.792 \emph{(0.855)} & 0.779 \emph{(0.812)} & 1.123 \emph{(1.475)} & 0.869 \emph{(0.644)} \\
Llama-3-Base & 0.847 \emph{(0.839)} & 1.182 \emph{(1.179)} & 1.472 \emph{(1.463)} & 1.297 \emph{(1.243)} \\
Gemma-2-Instruct & 0.760 \emph{(0.747)} & 1.014 \emph{(0.992)} & 0.996 \emph{(0.964)} & 1.176 \emph{(1.181)} \\
\bottomrule
\end{tabular}
\end{table}

\begin{figure}[t]
\centering
\includegraphics[width=0.5\textwidth]{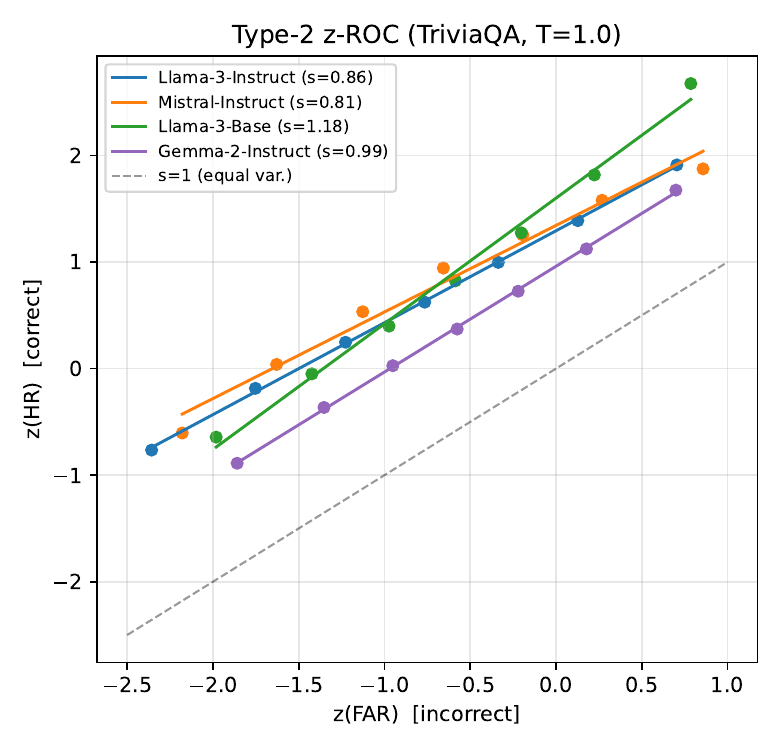}
\caption{Type-2 z-ROC (TriviaQA, $T{=}1.0$), corrected labels. Points are empirical (hit, false-alarm) pairs in probit coordinates; lines are SDT fits ($R^2 = 0.988$--$1.000$). Slopes below 1 (Mistral 0.78, Llama-3-Instruct 0.91) indicate greater variance in correct-answer evidence; Llama-3-Base is above 1 (1.18) and Gemma-2 near equal variance (1.01). The ordering replicates on Natural Questions.}
\label{fig:zroc}
\end{figure}

Table~\ref{tab:sdt} and Figure~\ref{fig:zroc} report the Type-2 SDT structure. The confidence signal has pronounced, model-specific unequal-variance structure. Mistral and Llama-3-Instruct have z-ROC slopes below 1 (0.81, 0.86; bootstrap CIs exclude 1), indicating the correct-answer evidence distribution is more variable than the incorrect. Llama-3-Base has the opposite structure ($s=1.18$, CI excludes 1) and Gemma-2 is close to equal variance ($s=0.99$). The ordering of slopes replicates on NQ (Table~\ref{tab:sdt}, final column). This structure is a genuine property of how each model represents correct versus incorrect answers, and is invisible to ECE and to AUROC$_2$ alone.

\subsection{Metacognitive information varies across models and co-varies with accuracy (H1, H4)}

\begin{table}[t]
\centering
\caption{Aggregate metacognition at $T{=}1.0$, \textbf{corrected labels}, ordered by accuracy. meta-I$_{2r}$ is bias-corrected (permutation null subtracted); 95\% bootstrap CIs from 2{,}000 trial-level resamples. All meta-I$_{2r}$ exceed the permutation null ($p<0.001$). Values reported in v2 are given in parentheses; see the version note for why they differ.}
\label{tab:agg}
\small
\begin{tabular}{lccccc}
\toprule
\textbf{Model} & \textbf{Acc} & \textbf{meta-I$_{2r}$} & \textbf{95\% CI} & \textbf{AUROC$_2$} & \textbf{meta-I$_{2r}$ (NQ)} \\
\midrule
Llama-3-Base & 0.478 \emph{(0.428)} & 0.300 \emph{(0.292)} & [0.282, 0.322] & 0.847 & 0.145 \emph{(0.140)} \\
Mistral-Instruct & 0.560 \emph{(0.427)} & 0.206 \emph{(0.328)} & [0.189, 0.226] & 0.792 & 0.085 \emph{(0.243)} \\
Llama-3-Instruct & 0.584 \emph{(0.543)} & 0.263 \emph{(0.276)} & [0.244, 0.284] & 0.833 & 0.157 \emph{(0.192)} \\
Gemma-2-Instruct & 0.627 \emph{(0.600)} & 0.151 \emph{(0.143)} & [0.135, 0.170] & 0.760 & 0.108 \emph{(0.108)} \\
\bottomrule
\end{tabular}
\end{table}

Metacognitive information varies by a factor of 1.98 across models
(meta-$I_{2r}$ 0.151--0.300 on TriviaQA; Table~\ref{tab:agg}). All four
estimates exceed their permutation null ($p<0.001$).

In v2, computed on the uncorrected labels, this section reported that the least
accurate model (Mistral) had the most informative confidence and the most
accurate (Gemma-2) the least, with meta-$I_{2r}$ and accuracy perfectly
inversely rank-ordered ($\rho={-}1.00$). That finding does not survive label
correction. Mistral's accuracy was understated by 13.3 points, and on corrected
labels it falls from the highest meta-$I_{2r}$ (0.328) to the third of four
(0.206). The ordering is now Llama-3-Base $>$ Llama-3-Instruct $>$ Mistral $>$
Gemma-2. The rank correlation with accuracy weakens to $\rho={-}0.80$ on
TriviaQA and vanishes on Natural Questions ($\rho={+}0.00$), where the ordering
is different again.

What remains is that cross-model variation in confidence informativeness is
substantial and permutation-significant, that Gemma-2 is reliably the least
informative on both datasets, and that the ordering is not predicted by
accuracy. We no longer claim an inverse coupling. The caveat raised in v2 ---
that a more accurate model's residual errors are harder near-misses and
therefore intrinsically less detectable --- is not needed to explain an effect
that is largely absent once the labels are corrected; the apparent coupling was
substantially an artefact of a scorer that penalised the most verbose model.

\begin{figure}[t]
\centering
\includegraphics[width=\textwidth]{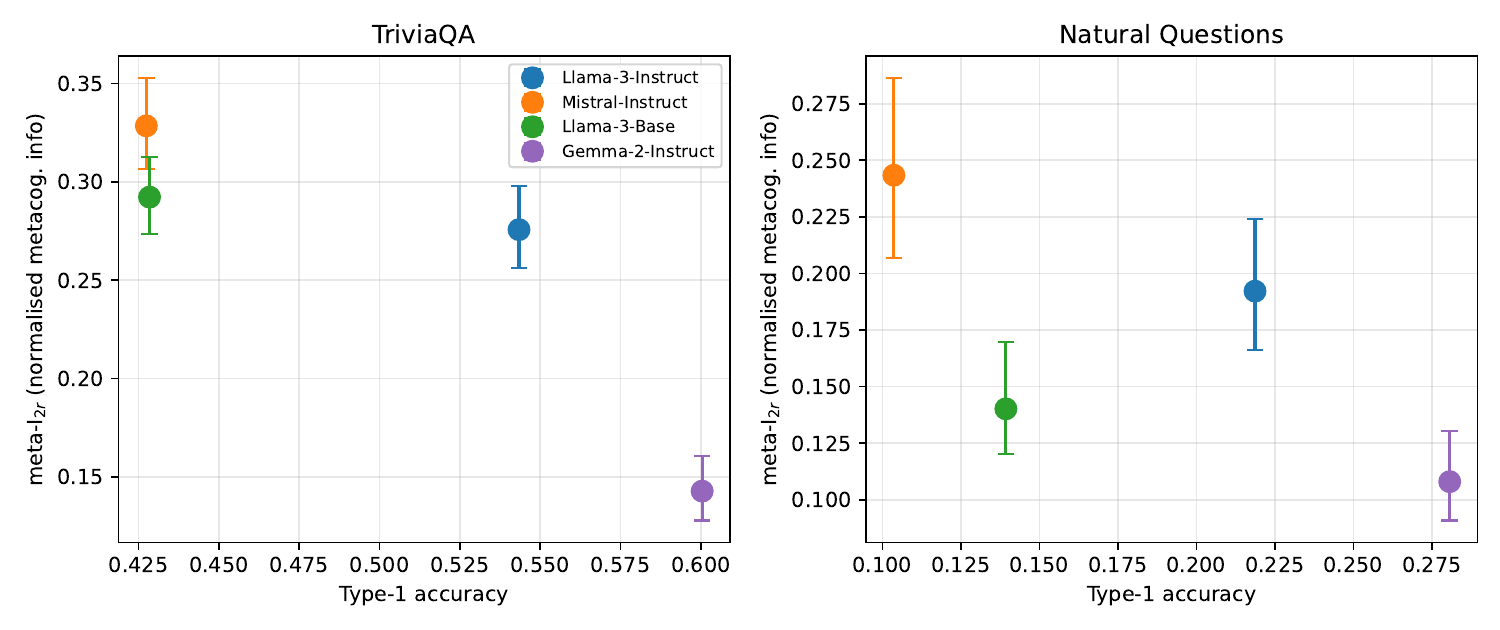}
\caption{Metacognitive information vs.\ accuracy, both datasets, corrected
labels. Error bars are 95\% bootstrap CIs. The monotone inverse relationship
reported in v2 is not present: the rank correlation is ${-}0.80$ on TriviaQA
and ${+}0.00$ on Natural Questions.}
\label{fig:metai2r}
\end{figure}

meta-$I_{2r}$ remains rank-identical to AUROC$_2$ on both datasets
($\rho={+}1.00$). The explanation offered in v2 --- that the four base-rate
entropies fall within 2.4\% of one another --- no longer holds on corrected
labels, where the spread is 4.6\% on TriviaQA and 23.3\% on NQ. The two
measures coincide here despite that spread; we report meta-$I_{2r}$ because it
is the model-free efficiency measure defined where meta-$d'$ is not
(\S\ref{sec:whynotmetad}), and note that AUROC$_2$ would support the same
conclusions on this sample.

\subsection{Domain-specific metacognitive information (H2)}

\begin{table}[t]
\centering
\caption{Domain-specific meta-$I_{2r}$ at $T{=}1.0$ on TriviaQA, \textbf{corrected labels}, all six classified domains. Boldface: weakest domain per model; \underline{underlined}: strongest. v2 reported only the four domains above the rule and stated that Arts \& Literature was strongest for every model; that holds within those four but not once Pop Culture and Sports are included.}
\label{tab:domain}
\small
\begin{tabular}{lcccc}
\toprule
\textbf{Domain} & \textbf{Llama-Inst} & \textbf{Mistral} & \textbf{Base} & \textbf{Gemma} \\
\midrule
History \& Politics & 0.244 & 0.192 & 0.297 & 0.143 \\
Arts \& Literature & 0.314 & 0.213 & 0.324 & 0.191 \\
Geography & 0.241 & 0.184 & 0.284 & 0.084 \\
Science \& Technology & \textbf{0.140} & \textbf{0.148} & \textbf{0.256} & \textbf{0.079} \\
\midrule
Pop Culture \& Ent. & 0.345 & \underline{0.303} & \underline{0.331} & \underline{0.275} \\
Sports & \underline{0.350} & 0.213 & 0.269 & 0.127 \\
\bottomrule
\end{tabular}
\end{table}

\begin{figure}[t]
\centering
\includegraphics[width=\textwidth]{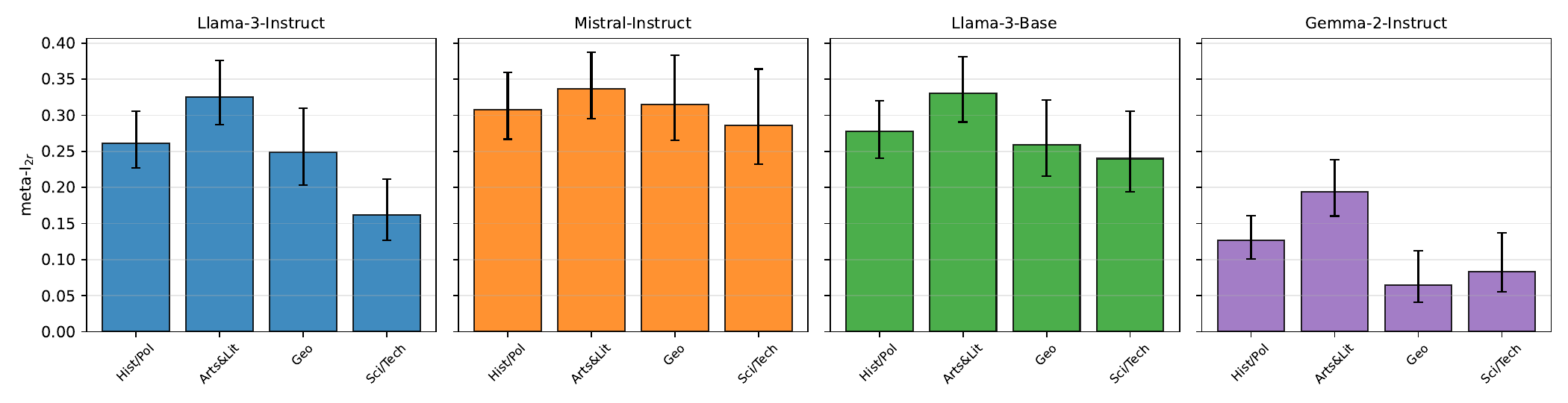}
\caption{Domain-specific meta-$I_{2r}$ at $T{=}1.0$ on TriviaQA, corrected labels, all six classified domains (95\% bootstrap CIs). Science \& Technology is the weakest domain for every model. The strongest is Pop Culture \& Entertainment for three models and Sports for Llama-3-Instruct; v2 reported Arts \& Literature as strongest, which holds only among the four domains it tabulated.}
\label{fig:domain}
\end{figure}

Metacognitive information varies systematically across domains within each model (Table~\ref{tab:domain}, Figure~\ref{fig:domain}). Science \& Technology is the weakest domain for all four models on corrected labels --- a stronger and more uniform result than v2 reported, which had Geography weakest for Gemma-2. The strongest domain is Pop Culture \& Entertainment for three models and Sports for Llama-3-Instruct. v2 stated that Arts \& Literature was strongest for every model; that holds among the four domains v2 tabulated, but not once Pop Culture and Sports are included, and we correct the claim here. Domain ranges remain model-dependent (Gemma-2 spans 0.079--0.275, Llama-3-Base 0.256--0.331). A model whose confidence is informative in one domain and uninformative in another poses a deployment risk that aggregate metrics conceal.

\subsection{Temperature dissociates accuracy from metacognitive information (H3)}

\begin{figure}[t]
\centering
\includegraphics[width=\textwidth]{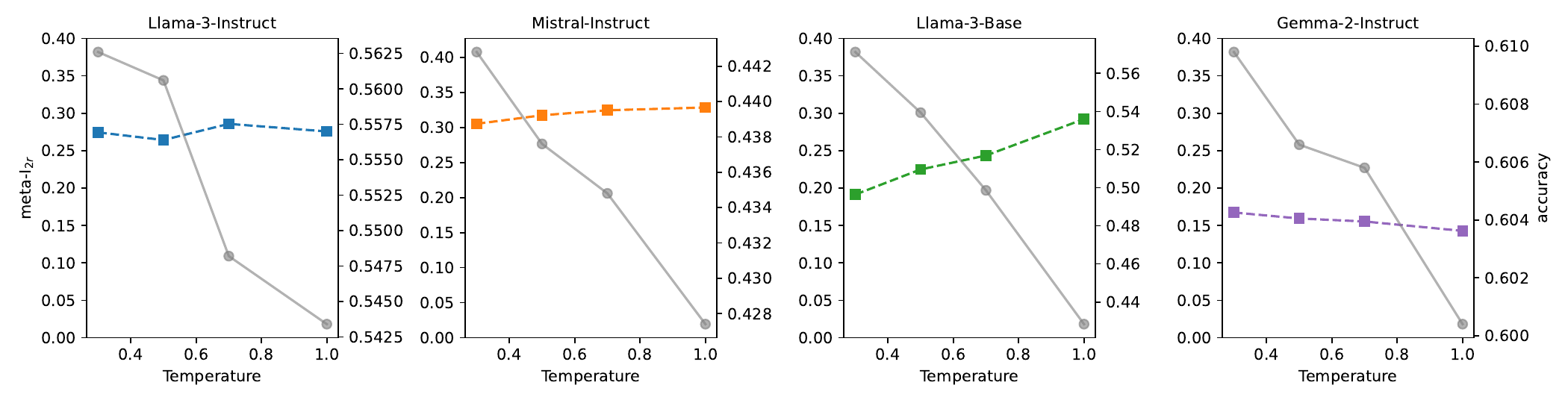}
\caption{meta-$I_{2r}$ (coloured, dashed) and accuracy (grey, solid) vs.\ temperature on TriviaQA, corrected labels. Accuracy falls with temperature for all four models, monotonically for Llama-3-Base and Llama-3-Instruct and near-monotonically for the other two; metacognitive information is near-flat for three of four, dissociating the two. Llama-3-Base is the exception, its confidence becoming markedly more informative as accuracy falls.}
\label{fig:temperature}
\end{figure}

Temperature dissociates Type-1 accuracy from metacognitive information (Figure~\ref{fig:temperature}). Accuracy decreases with temperature for all four models, monotonically for
Llama-3-Base and Llama-3-Instruct ($\rho(\text{acc},T)=-1.00$) and near-monotonically
for Gemma-2 ($-0.93$) and Mistral ($-0.82$); v2 reported $-1.00$ for all four,
which does not hold on corrected labels. Metacognitive information is near-flat
for three of four (meta-$I_{2r}$ range 0.009 for Mistral, 0.016 for
Llama-3-Instruct and 0.027 for Gemma-2 across $T\in\{0.3,0.5,0.7,1.0\}$),
moving in a different direction from accuracy. Llama-3-Base is again the
exception (range 0.103), the only model whose confidence informativeness
changes materially with temperature. Temperature reshapes what a model gets right without, for most models, changing how well its confidence tracks correctness.

\subsection{Selective prediction (deployment consequence)}

The cross-model variation in confidence informativeness has a direct deployment consequence for selective prediction, where a system abstains on low-confidence responses. What metacognitive information predicts is the \emph{gain} from confidence-based selection, not the absolute accuracy attained---the latter is dominated by base accuracy. This is also where the accuracy confound above matters less: gain is measured within each model, so it is not driven by the cross-model accuracy differences that confound the aggregate ordering. At 50\% coverage (accepting the top half of responses ranked by NLP),
Llama-3-Base improves from 47.8\% to 74.0\% ($+26.3$ points) and
Llama-3-Instruct from 58.4\% to 82.8\% ($+24.4$), while Gemma-2, with the least
informative confidence, gains least (62.7\% to 80.1\%, $+17.4$). Across the
four models, meta-$I_{2r}$ tracks the selective-prediction gain exactly on
corrected labels (Spearman $\rho={+}1.00$, against $+0.80$ in v2) and not the
accuracy level ($\rho={-}0.40$), which follows base accuracy instead. This is the
one result that strengthens under correction. A model with high accuracy but low metacognitive information (Gemma-2) reaches a high accuracy level under selection yet extracts little additional value from its own confidence; a model with lower accuracy but informative confidence (Mistral) benefits most from abstention. Neither pattern is visible from accuracy or ECE alone (full accuracy--coverage curves in Appendix~\ref{app:selective}).

\section{Discussion}

\subsection{What an SDT analysis adds to LLM evaluation}

Confidence evaluation operates at three tiers. \emph{Tier~1} (ECE, Brier score) measures alignment, conflates sensitivity with bias, and is unstable under discretisation. \emph{Tier~2} (AUROC$_2$, rank correlations) measures ranking quality; it does not normalise for the base rate, though on this sample the two measures rank identically regardless. \emph{Tier~3} (meta-$I_{2r}$, and the z-ROC structure of the confidence signal) measures how informative the confidence signal is about correctness with a model-free estimator valid where meta-$d'$ is not. Our results make the practical consequence concrete: the confidence signals of different models have qualitatively different variance structure ($s$ from 0.78 to 1.18), and confidence informativeness varies by a factor of 1.98 across models---both invisible to ECE. On corrected labels the cross-model ordering is not predicted by accuracy (\S\ref{sec:results}); the contributions that survive relabelling are the demonstrated variance structure, the domain and temperature effects, and the selective-prediction result.

\subsection{Temperature, criterion, and metacognitive capacity}

The dissociation between temperature and metacognitive information has implications for temperature tuning. For three of four models, temperature moves Type-1 accuracy without changing how well confidence tracks correctness. If a model's confidence is uninformative, lowering temperature to raise accuracy will not make its confidence more useful for selective prediction. For Llama-3-Base, where the dissociation does not hold, temperature does change the information content of confidence itself.

\subsection{Connections to human metacognition}

The phenomena parallel human metacognition, where metacognitive efficiency is domain-specific~\citep{rouault2018domain}, dissociable from Type-1 performance~\citep{fleming2010relating}, and neurally distinct from perceptual decisions~\citep{fleming2012prefrontal}. Our findings parallel these results functionally, not mechanistically: we do not claim LLMs possess metacognition phenomenologically. The measure differs from the human meta-$d'$ literature because open-ended QA lacks a Type-1 decision (\S\ref{sec:whynotmetad}); the value lies in the decomposition, consistent with the use of SDT in medical diagnosis~\citep{swets1996signal} and automated system evaluation~\citep{bartlett2017evaluating}.

\subsection{Limitations}

\paragraph{Residual scoring error.} The containment correction applied in this
version recovers roughly half the gap to human accuracy and halves the
cross-model differential, but does not remove it: adjudicating a further 630
blinded responses across seven models, 18.4\% of trials the corrected scorer
still rejects are in fact correct, and this residual rate remains
model-dependent by a factor of 2.4. The values reported here are therefore
better than v2's but should not be treated as final, and fine-grained
cross-model orderings should be read with that in mind.

\label{sec:limitations}

Four open-weight 7--9B models; generalisability to frontier scale is unknown. API models that do not expose token-level log-probabilities cannot be evaluated with internal NLP; the verbal-confidence approach of \citet{dai2026rescaling} is a complementary path. All models ran as Q5\_K\_M quantisations; however, all measures depend on the \emph{ordinal} relationship between NLP and accuracy, which quantisation preserves, and the monotonicity check confirms this in all conditions. meta-$I_{2r}$ belongs to a family of information-theoretic metacognition measures~\citep{dayan2023metacognitive, fitousi2025information} with several normalisation choices; we use normalisation by accuracy entropy and report permutation nulls throughout to guard against the small-sample bias of plug-in mutual information. NLP is a fluency measure, not a pure metacognitive signal; a high meta-$I_{2r}$ does not imply the model ``knows that it knows'' in any deep sense.

\subsection{Recommendations for practice}
\begin{enumerate}[leftmargin=*, nosep]
    \item Report meta-$I_{2r}$ (with a permutation null) alongside ECE; they are complementary.
    \item Report the z-ROC slope of the confidence signal: unequal variance is a real, model-specific property.
    \item Disaggregate by domain; aggregate metrics hide domain-specific metacognitive deficits.
    \item For confidence-dependent systems, prefer higher metacognitive information over lower ECE.
    \item Evaluate temperature effects on metacognitive information, not just calibration.
\end{enumerate}

\section{Conclusion}

We have applied Signal Detection Theory to LLM confidence, characterising the Type-2 ROC of the confidence signal and its unequal-variance structure, and---because the meta-$d'$ efficiency ratio is undefined for open-ended QA---quantifying metacognitive efficiency with a model-free information measure, meta-$I_{2r}$. Across four models and 224{,}000 trials, metacognitive information varies by a
factor of 1.98 and is not predicted by accuracy: the inverse coupling reported
in earlier versions of this paper was substantially an artefact of a
differentially length-biased correctness scorer, and does not survive
relabelling. The confidence signal has model-specific variance structure, efficiency is domain-specific, and temperature dissociates accuracy from metacognitive information. Current practice treats confidence as monolithic; an SDT decomposition shows this is insufficient. All analyses are pre-registered, with code and data publicly available.\footnote{Pre-registration: \url{https://osf.io/5q7mt}. Code and data:
\url{https://github.com/synthiumjp/sdt_calibration}. The corrected scorer, the
1{,}830 human adjudications used to validate it, and the recomputation scripts
for this version: \url{https://github.com/synthiumjp/metacognition-audit}.}

\paragraph{Use of Generative AI.} Claude (Anthropic) was used as a research assistant for analysis pipeline design and code generation. All scientific decisions, hypothesis formulation, and interpretive judgments were made by the author.

\bibliographystyle{plainnat}
\bibliography{references}

\appendix

\section{NLP Monotonicity Check}
\label{app:monotonicity}
Accuracy increases strictly across NLP quartiles in all eight model$\times$dataset conditions at $T{=}1.0$ (e.g., Gemma-2 on TriviaQA: $Q_1{=}0.314$, $Q_2{=}0.539$, $Q_3{=}0.690$, $Q_4{=}0.859$), validating NLP as a graded evidence variable.

\section{Selective Prediction: Accuracy--Coverage}
\label{app:selective}
Accuracy as a function of coverage (fraction of queries answered), abstaining on lowest-confidence responses. Models with higher meta-$I_{2r}$ obtain a larger accuracy \emph{gain} from confidence-based abstention (Spearman $\rho={+}0.80$ between meta-$I_{2r}$ and the accuracy gain at 50\% coverage), while the absolute accuracy level under selection is governed by base accuracy. This confirms that metacognitive information predicts the value extracted from a model's own confidence signal, not the accuracy ceiling.

\end{document}